\documentclass[runningheads]{llncs}

\usepackage{eccvabbrv}

\usepackage{graphicx}
\usepackage{booktabs}

\usepackage[accsupp]{axessibility}  

\usepackage{amsthm}
\usepackage[pagebackref,breaklinks,colorlinks]{hyperref}

\usepackage{orcidlink}
\usepackage{algorithm}
\usepackage{algorithmic}
\usepackage{booktabs} 
\usepackage{multirow}
\usepackage{amsmath}
\usepackage{amssymb}
\usepackage{mathtools}

\usepackage{stfloats}
\usepackage{array,multirow,graphicx}

\usepackage{newfloat}
\usepackage{listings}

\begin{document}

\title{Adversarial Training via Adaptive Knowledge Amalgamation of an Ensemble of Teachers} 

\titlerunning{Abbreviated paper title}

\author{Shayan Mohajer Hamidi\inst{1}\orcidlink{0000-0001-8321-7130} \and
Linfeng Ye\inst{1}}
\institute{Department of Electrical and Computer Engineering University of Waterloo, Canada.\\
\email{\{smohajer,l44ye\}@uwaterloo.ca }}

\maketitle

\begin{abstract}
Adversarial training (AT) is a popular method for training robust deep neural networks (DNNs) against adversarial attacks. Yet, AT suffers from two shortcomings: (i) the robustness of DNNs trained by AT is highly intertwined with the size of the DNNs, posing challenges in achieving robustness in smaller models; and (ii) the adversarial samples employed during the AT process exhibit poor generalization, leaving DNNs vulnerable to unforeseen attack types. To address these dual challenges, this paper introduces \textbf{a}dversarial \textbf{t}raining via \textbf{a}daptive \textbf{k}nowledge \textbf{a}malgamation of an ensemble of teachers (AT-AKA). In particular, we generate a diverse set of adversarial samples as the inputs to an ensemble of teachers; and then, we adaptively amalgamate the logtis of these teachers to train a generalized-robust student. Through comprehensive experiments, we illustrate the superior efficacy of AT-AKA over existing AT methods and adversarial robustness distillation techniques against cutting-edge attacks, including AutoAttack. 
\end{abstract}

\section{Introduction} \label{Sec:intro}
In recent years, deploying deep neural networks (DNNs) in various fields such as computer vision and neural language processing has gained considerable momentum. Yet, DNNs are vulnerable to adversarial samples crafted
by adding small and human-imperceptible adversarial perturbations to normal examples \cite{goodfellow2014explaining,sun2021watermarking}. In particular,
safety and security focused applications, for instance face recognition \cite{kurakin2018adversarial} and autonomous driving \cite{chen2015deepdriving}, are concerned about robustness to adversarial samples. To this end, many recent works have aimed to train \textit{robust} DNNs which can achieve a certain level of accuracy not only on clean samples, but also on adversarial samples
\cite{goodfellow2014explaining,kurakin2016adversarial,liao2018defense,wong2018provable,pang2018towards,xie2019feature,mohajer2024robustness,ye2021thundernna,zhang2019theoretically,10446776}.

One of the proposed defense strategies against adversarial attacks is adversarial training (AT) \cite{goodfellow2014explaining,madry2017towards}
which is shown to be effective against
adversarial attacks \cite{pang2020bag,maini2020adversarial,schott2018towards} to some extent. The idea of AT is to replace the training samples with their adversarial
versions in each training loop in the hope that
an adversarially trained model behaves normally when it is fed by adversarial
samples. Specifically, \cite{madry2017towards} formulated AT as a min-max problem,
searching for the best solution to the worst-case optimum. Since the inner maximization problem in AT is non-concave and typically intractable, it is often approximated by fast gradient sign method (FGSM) \cite{goodfellow2014explaining} or projected gradient descent (PGD) method \cite{madry2017towards}. In general, AT faces two daunting challenges:

\noindent \textit{Challenge (I)}: The robustness of models trained within the AT framework primarily hinges on model capacity, with larger models generally exhibiting greater robustness \cite{zi2021revisiting,wu2021wider}. Nevertheless, the adoption of large models is impractical for resource-constrained devices due to their substantial storage overhead, as well as their demand for high computation and memory resources.

\noindent \textit{Challenge (II)}: Adversarial samples generated by attack methods in AT often exhibit poor generalization to unseen attacks \cite{song2018improving}. This limitation arises because a single attack algorithm can only produce one type of adversarial sample. Indeed, an example of this limitation can be observed in the PGD attack, where adversarial samples, even with random initialization, often result in nearly identical loss values \cite{madry2017towards}, highlighting a lack of diversity \cite{tashiro2020diversity}. Consequently, DNNs trained with a limited range of adversarial examples fail to encounter diverse versions of adversarial samples, making them vulnerable to various types of attacks.

To address \textit{Challenge (I)}, an effective approach is adversarial robustness distillation. This method entails transferring the robustness acquired by a large model (referred to as the teacher) trained through AT to a smaller model (known as the student) \cite{zi2021revisiting,shu2021encoding,goldblum2020adversarially,chen2021robust}. In particular, this process involves mimicking the soft labels of the adversarially pretrained teachers, thus transferring their robustness to the student models. Yet, while these methods have shown success in training robust models with small sizes, they still fail to address \textit{Challenge (II)}. 

On the contrary, to address \textit{Challenge (II)}, earlier studies have utilized an ensemble of sub-models in generating diverse adversarial examples \cite{yang2021trs,yang2020dverge,pang2019improving}. While the intention behind these approaches is to enhance sub-model diversity, they face two main challenges. Firstly, empirical results indicate that their utilized diversity metrics are ineffective in inducing output diversity among sub-models; consequently, the corresponding ensemble struggles to achieve the desired level of robustness \cite{tramer2020adaptive}. Secondly, they still encounter \textit{Challenge (I)}, wherein the issue of robustness in smaller models persists.

In this paper, to effectively tackle \textit{Challenges (I)} \& \textit{(II)} simultaneously, we propose \textbf{a}dversarial \textbf{t}raining via \textbf{a}daptive \textbf{k}nowledge \textbf{a}malgamation of an ensemble of teachers (AT-AKA). In AT-AKA, we distill robust knowledge from an ensemble of teachers (sub-models) into a lightweight student model. Within this framework, each teacher undergoes training via AT while it is fed with a distinctive version of adversarial samples derived from benign inputs. As such, each teacher becomes robust against a distinct version of adversarial sample. This diversity among the teachers facilitates the distillation of a more generalized robustness knowledge into the student model. To achieve this diversity in adversarial samples, we employ Stein variational gradient descent (SVGD) \cite{liu2016stein}. Notably, the update rule of SVGD comprises two terms: the first term functions similarly to FGSM and PGD, focusing on maximizing the loss value, while the second term acts as a repulsive force to encourage sample diversity. Following the generation of these diverse adversarial samples, we propose various adaptive amalgamation methods to fuse the knowledge of these teachers. These methods are designed (i) with consideration of the inner-maximization problem in AT and (ii) in an adaptive manner, wherein we combine the logits of teachers based on their respective loss values. This comprehensive approach aims to equip the student model with robust classification boundaries effective against a variety of attacks. In summary, the contributions of this paper are as follows:
\begin{itemize}
\item We employ an ensemble of teachers to train a compact yet adversarially robust student model, a framework we term AT-AKA. Within this framework, the teachers are exposed to a diverse array of adversarial samples. This diversity serves the purpose of assisting the student in acquiring an enhanced understanding and generalization of adversarial samples.

\item To generate such diverse/distinguished adversarial samples, we exploit SVGD algorithm. 

\item We deploy a \textit{dynamic} sample adaptive weighting strategy to adaptively distill knowledge from the ensemble of teachers.

\item We further incorporate AT-AKA into collaborative KD \cite{guo2020online}, yielding collaborative AT-AKA (CAT-AKA). 

\item By conducting thorough experiments, compared to the state-of-the-art algorithms, we show that the proposed method can yield a high adversarial robustness while maintaining a high clean accuracy. 
\end{itemize}
\section{Related Works}
\label{Sec:Related Work}
\subsection{Adversarial defenses}
Considerable efforts have been put into devising methods to make the DNNs robust against adversarial attacks, such as input de-noising \cite{song2017pixeldefend} or
feature de-noising \cite{xie2019feature} (more methods can be found in \cite{kurakin2018adversarial}). Among these methods, adversarial training (AT) is
shown to be one of the most effective defensive methods against adversarial attacks (AT was first introduced by \cite{szegedy2013intriguing,goodfellow2014explaining} and then theoretically studied and justified by \cite{madry2017towards}).
AT aims to incorporate the adversarial search within the training process, and consequently to realize robustness against adversarial examples during the test stage. Since \cite{madry2017towards}, some works in the literature mentioned the weaknesses of this framework. For example, \cite{zhang2019limitations} showed that adversarialy trained
models are vulnerable to ‘blind-spot’ attacks. In addition, many works discussed that the adversarial training yields a poor \emph{generalized} robustness, in that these networks are not robust against all different types of attacks \cite{song2018improving,geirhos2018imagenet,zhang2019interpreting,gong2021maxup}. In fact, the aim of this work is to alleviate this issue, and to train a \emph{generalized} robust model.

\subsection{Adversarial training via knowledge distillation}
Knowledge distillation (KD) is a framework via which knowledge from a pre-trained cumbersome model (teacher) is distilled into a smaller model (student) \cite{hinton2015distilling}. KD has been widely exploited in many applications such as image recognition, semantic segmentation, and especially model compression \cite{polino2018model,tung2019similarity,wang2021knowledge}. In KD, the student model mimics the prediction of pre-trained teacher model to make itself more powerful than it’s trained alone. Knowledge distillation could be carried out in offline or online \cite{chen2020online,guo2020online,wu2021peer,bhat2021distill,10487854} fashions. 

The offline KD deploys a two-stage training mode, in that it first trains the teacher model, and then trains the student network by some  distillation strategies \cite{hinton2015distilling}. Yet, the two-stage nature of offline KD will increases both training cost and pipeline complexity. On the other hand, online KD eliminates the necessity of
pre-training a cumbersome model, and lets all models be trained at the same time in one stage.

Inspired by KD, \cite{papernot2016distillation} introduced the notion of transferring adversarial robustness, referred to as defensive distillation. This method aims to tackle the issue with AT that it generally yields better robustness for larger models \cite{rice2020overfitting,zi2021revisiting}. This method requires the student and teacher models have the same architectures, and was shown that it does not make the decision boundary secure and it is not robust against general attacks \cite{bai2023guided}. In \cite{goldblum2020adversarially}, authors suggested that using a cumbersome teacher model in adversarial training allows better adversarial training strategies. 
\cite{zhu2021reliable} deployed the teacher model with the same structure as the student model for adversarial knowledge distillation. The work in \cite{zi2021revisiting} exploited teacher's soft label (instead of the one-hot label) in order to generate adversarial examples in the process of adversarial training which yields improvement in the robustness of the student model.

\subsection{Ensemble of sub-models for robustness}
Given the tremendous success of ensemble methods, researchers have recently been exploring ways to enhance the robustness of an ensemble consisting of small sub-models. 

Several studies tried to encourage diversity in internal representations or outputs across sub-models, as a means to curtail adversarial transferability and enhance the overall robustness of the ensemble. By promoting diversity within the ensemble, the vulnerability to adversarial attacks can be minimized, leading to more reliable and secure results. For instance, \cite{pang2019improving} forces different sub-models to have high diversity in the non-maximal predictions. \cite{kariyappa2019improving} maximizes the cosine distance between
each sub-model’s gradient w.r.t. the input to reduce the overlap between
adversarial subspaces for different sub-models.

\section{Preliminaries and Notations}
\label{Sec:notations}
\subsection{Notations}
We denote by $[n]$ the set of integers $\{1,2,\cdots,n\}$. In addition, we define $\{x_k\}_{k \in [K]}=\{x_1,x_2,\dots,x_K\}$ for a scalar/vector $x$. Scalars are denoted by lowercase letters (e.g., $x$), and vectors are represented by bold-face lowercase letters (e.g., $\boldsymbol{x}$). We use $\|\boldsymbol{x}\|_p$ to show the $l_p$ norm of vector $\boldsymbol{x}$. Denote by $\pi_\mathcal{C}(\boldsymbol{x})$ the projection of vector $\boldsymbol{x}$ onto closed set $\mathcal{C}$; that is, $\pi_\mathcal{C}(\boldsymbol{x})=\arg \min_{\boldsymbol{y}}\{\| \boldsymbol{x} -\boldsymbol{y}\|_2 ~|~ y \in \mathcal{C}\}$. Furthermore, we denote by $S$ and $\{T_i\}_{i \in [n]}$ the student model and the set of $n$ teacher models, respectively.

\subsection{Preliminaries}
\noindent$\bullet$ \textbf{Knowledge distillation.}
Knowledge distillation (KD) \cite{hinton2015distilling} acts as a complementary method to further improve
the performance of a small model (student) by distilling the knowledge from a cumbersome model (teacher). In fact, during the training phase, the student deploys an extra supervision provided by the teacher in conjunction with its conventional supervised learning objective such as the cross-entropy loss.

As such, the student is encouraged to mimic teacher’s behavior either (i) by minimizing the Kullback-Leibler divergence of predictions \cite{hinton2015distilling}, or (ii) by minimizing the euclidean distance of feature representations \cite{li2017mimicking} between teacher and student.

\noindent$\bullet$ \textbf{Stein variational gradient descent.}
In Bayesian inference, subjective probabilities, so-called prior distribution, are assigned to the distributions in order to generate some data. After observing the data, Bayes' rule is used to update the prior to posterior distribution. Yet, computing posterior distributions are often intractable. As such, Markov chain Monte Carlo (MCMC) has been widely used in probabilistic inference to draw approximate posterior
samples; however, it is often time-consuming and faces difficulties in convergence \cite{liu2016stein}. To aleviate this problem, \cite{liu2016stein} proposed Stein variational gradient descent (SVGD) which is a stochastic variational particle-based approach. In fact, by maintaining a flow of distributions, SVGD provides a solid theoretical guarantee
of the convergence of the set of particles to the target distribution. In particular, SVGD starts from an easy-to-sample initial (prior) distribution and learns
the subsequent (posterior) distribution in the flow by push-forwarding the current one using a function $T(x)=x+\eta \phi (x)$, where $x$ is a sample drawn from some distribution, $\eta>0$ is a small step-size, and $\phi(\cdot)$ is a
nonlinear function described by a reproducing kernel Hilbert space with given kernel $k(x,T(x))$. 

\noindent$\bullet$ \textbf{Adversarial training.}
As formulated in \cite{madry2017towards}, the goal of AT is to solve the following optimization problem
\begin{align} \label{eq:AT}
\boldsymbol{\theta}^*=\arg \min_{\boldsymbol{\theta}} \mathbb{E}_{(\boldsymbol{x},y) \sim \mathcal{D}} \{ \max_{\| \boldsymbol{\epsilon}\|_p<\epsilon_{\text{max}}} \mathcal{L}(f(\boldsymbol{x}+\boldsymbol{\epsilon};\boldsymbol{\theta}),y)\} ,  
\end{align}
where $\mathcal{D}$ is the training samples, and $\boldsymbol{\theta}$ represents the model parameters. Solving the inner-maximization in Eq. \eqref{eq:AT} has been known to be a challenging task in the literature. 

\begin{figure*}
\centering
\includegraphics[width=0.9\textwidth]{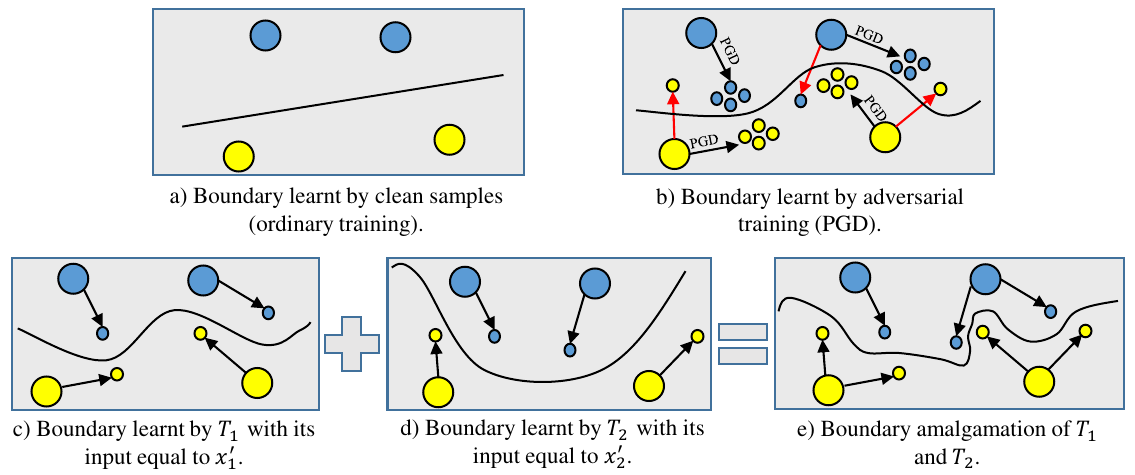}
  \caption{A two-dimensional caricature of the classification boundaries learnt by different methods: (a) an ordinary learning method using the clean samples; (b) adversarial training where PGD is used to generate adversarial samples; (c) $T_1$ via first set of adversarial samples; (d) $T_2$ via second set of adversarial samples; and (e) amalgamation of $T_1$ and $T_2$.} \label{fig:motiv}
  \vskip -0.1in
\end{figure*} 
\section{Motivation} \label{Sec:Motivation}
A popular method to train a robust neural network is AT, where the model is fed with adversarial versions of training samples during the training stage. These adversarial samples are generated by some attack methods, such as FGSM or PGD, which often yield poorly-diversified attacked versions of clean samples \cite{tashiro2020diversity}. An illustration of attacked samples generated by PGD attack is depicted in Fig. \ref{fig:motiv} (b) where the PGD-attacked samples corresponding to each clean sample are concentrated in some specific area. Consequently, the classification boundaries learnt by these samples during adversarial training are formed to only correctly classify these samples. Therefore, an adversarially-trained model via these concentrated samples shows a poor robustness when facing new attacked samples. For instance, some samples against which the trained model is not robust are depicted in Fig. \ref{fig:motiv} (b) (the misclassified samples are distinguished using red arrows).  Similarly, this problem is also existing in the AT methods deploying KD since they use the same attack strategies in order to generate adversarial samples. 

To tackle this issue, in this paper, we propose the following remedy: First, we generate a diverse set of adversarial samples by deploying SVGD; then, we feed each of these distinct samples into distinct teacher models such that each teacher learns a classification boundary for its respective adversarial sample. Lastly, we combine these boundaries learnt by different teachers and distill this into the student via logits matching. A caricature of such boundary learning strategy is depicted in Figs. \ref{fig:motiv} (c-e) for the case of having two teachers. Specifically, the boundaries learnt by $T_1$ and $T_2$ are depicted in Fig. \ref{fig:motiv} (c) and Fig. \ref{fig:motiv} (d), respectively; and the amalgamation of these boundaries are depicted in Fig. \ref{fig:motiv} (e). As seen in Fig. \ref{fig:motiv} (e), the model is now robust against a variety versions of attacks. In the following section, we elaborate on the methodology of AT-AKA. 

\section{Methodology and Formulation
} \label{Sec:Formulation}
\begin{figure}[!t]
\centering\includegraphics[width=0.7\linewidth]{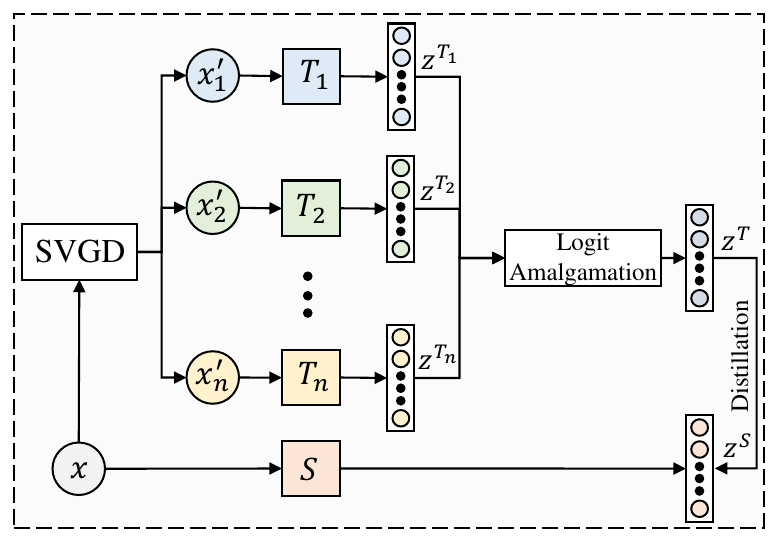}
    \caption{The AT-AKA framework. The teachers and the student are denoted by $\{T_i\}_{i \in [n]}$ and $S$, respectively. 
    All the $\{T_i\}_{i \in [n]}$ and $S$ are trained from scratch. The clean sample $\boldsymbol{x}$ is fed to the student, however, $n$ distinct adversarial samples generated by SVGD are fed to the teachers. Then, the logits of the teachers are amalgamated, and the resulting logit is used to distill knowledge into the student.} 
\vskip -0.1in
    \label{fig:AT-AKA}
\end{figure}
Fig. \ref{fig:AT-AKA} depicts the mechanism of AT-AKA where the student model is fed with clean sample $\boldsymbol{x}$, and the teachers $\{T_i\}_{i \in [n]}$ are fed with adversarial examples $\{\boldsymbol{x}^{\prime}_i \}_{i \in [n]}$ generated from $\boldsymbol{x}$. First, in subsection \ref{sec:svgd}, we discuss how we generate a diverse set of adversarial samples $\{\boldsymbol{x}^{\prime}_i \}_{i \in [n]}$. Afterward, we explain the mechanism of the AT-AKA.

\subsection{Generating adversarial samples} \label{sec:svgd}
To generate $\{\boldsymbol{x}^{\prime}_i \}_{i \in [n]}$, the clean sample $\boldsymbol{x}$ goes through the SVGD block whose mechanism is explained in the sequel. 

Similarly to \cite{doan2022bayesian,nguyen2023adversarial}, we use SVGD to generate a set of adversarial samples $\{\boldsymbol{x}^{\prime}_i \}_{i \in [n]}$. To shed more light, first, we assume a prior distribution $p(\boldsymbol{x})$ for $\{\boldsymbol{x}^{\prime}_i \}_{i \in [n]}$ (we use uniform prior distribution in our experiments). Then we iteratively update these points (points/samples are referred to as particles in the SVGD method) such that after enough number of iterations, the empirical distribution of these samples tends to that of adversarial samples for $\boldsymbol{x}$.  Denote by $\{\boldsymbol{x}^{\prime}_{i,t} \}_{i \in [n]}$ the set of points generated at the $t$-th iteration of the SVGD algorithm. Hence, $\{\boldsymbol{x}^{\prime}_{i,0} \}_{i \in [n]}=\{\boldsymbol{x} + \epsilon_i \}_{i \in [n]}$ is the input to the SVGD algorithm, where $\epsilon_i $ is a random variable drawn from uniform distribution $U(-\epsilon,\epsilon)$, and $\epsilon$ indicates the adversarial budget. At iteration $t+1$, the $i$-th point is updated as follows
\small
\begin{align} \label{eq:SVGD}
&\boldsymbol{x}^{\prime}_{i,t+1} =\pi_\mathcal{C} \left( \boldsymbol{x}^{\prime}_{i,t} +\eta \phi (\boldsymbol{x}^{\prime}_{i,t}) \right).  
\\ 
& \text{with} \qquad \phi(\boldsymbol{x})=\sum_{j=1}^n[k(\boldsymbol{x}^{\prime}_{j,t},\boldsymbol{x}) \nabla_{\boldsymbol{x}^{\prime}_{j,t}} \mathcal{L} (f_{T_i}(\boldsymbol{x}^{\prime}_{j,t}),y)  \frac{\gamma}{n}\nabla_{\boldsymbol{x}^{\prime}_{j,t}}k(\boldsymbol{x}^{\prime} _{j,t},\boldsymbol{x})], \nonumber
\end{align}
\normalsize
where $\eta$ is the step-size, $k(\cdot,\cdot)$ is a kernel function that measures the similarity between the adversarial points, $\gamma$ is a hyper-parameter. Note that the gradients in Eq. \eqref{eq:SVGD} are calculated for the teacher models. 

As seen in Eq. \eqref{eq:SVGD}, $\phi(\boldsymbol{x})$ has two terms: (i) the first term is the weighted sum of the gradients of all the
points weighted by the kernel function, and thus the nature of the first term is the same as that of gradient descent method; and (ii) the second term behaves as a repulsive force that prevents
all the adversarial points to collapse together, which enforces diversity among the obtained adversarial points. 

\subsubsection{SVGD Vs. PGD}
Now, we conduct the following experiments to show that SVGD can generate more diverse adversarial samples compared to those generated by PGD. To this end, we randomly pick a sample in CIFAR-10 dataset, and generate two sets of 10 attacked samples obtained by: 

\noindent $\bullet$ Set 1: 10 runs of PGD with each run starts at a uniformly random point in the allowed perturbation budget; 

\noindent $\bullet$ Set 2: SVGD with 10 initial particles.

For both SVGD and PGD, the number of iteration is equal 20 with the same step-size. Also, the perturbation budget is $\|\boldsymbol{\epsilon}\|_{\infty}<8/255$. Then, for these two sets, we measure both the strength and the spacial diversity of the attacked samples in these two sets. To measure how spatially-diversified the samples are, we measure the average pairwise distance ($l_{\infty}$ distance) between the samples in each set, denoted by $dist$. In addition, to see how strong the attacking method is, we measure the average, standard deviation and the maximum of cross-entropy (CE) loss values for the attacked samples (see \ref{table:svgdvspgd}). As seen, the adversarial samples generated by SVGD are far more diverse than those generated by PGD. In addition, SVGD is also able to generate a strong attack (max CE value), however, most of the adversarial samples generated by PGD are strong with almost the same value.

\begin{table}[t]
\caption{SVGD Vs. PGD}
\vskip -0.2in
\label{table:svgdvspgd}
\begin{center}
\begin{small}
\resizebox{0.7\columnwidth}{!}{%
\begin{tabular}{l||c|c|c|c}
\toprule
Set & $dist$ & Avg(CE) & Std(CE) & Max(CE) \\
\midrule
Set 1 (generated by PGD)  & 17.1 &  41.2 & 3.4 & 43.2 \\
Set 2 (generated by SVGD) & 78.4 &  37.8 & 7.5 & 43.7 \\
\bottomrule
\end{tabular}}
\end{small}
\end{center}
\vskip -0.3in
\end{table}

\begin{remark} \label{rem:RBF}
It is known that when deploying Gaussian
RBF kernel, by letting the kernel width approach $+\infty$, the update formula of SVGD at each step
asymptotically reduces to the typical gradient descent (GD). Hence, the update rule in Eq. \eqref{eq:SVGD} reduces to FGSM (or equivalently PGD).
\end{remark}

\subsection{AT-AKA mechanism} \label{sec:mechanism}
In AT-AKA, the teacher models and the student, are trained from scratch during the training phase\footnote{In this work, we abusively refer to the cumbersome models as "teachers". Since all the models are trained from scratch, one could refer to all the models as students.}. Therefore, AT-AKA falls into the category of \textit{online} KD. 
After the advesarial samples $\{\boldsymbol{x}^{\prime}_i \}_{i \in [n]}$ are generated, they are fed into teachers $\{T_i\}_{i \in [n]}$, yielding respective logic vectors $\{\boldsymbol{z}^{T_i}\}_{i \in [n]}$. In order to combine the knowledge of the teachers,
an amalgamation function $\psi(\cdot)$ is exploited whose input is an $n$-tuple logits vectors. We denote the logits vector obtained after combining $\{\boldsymbol{z}^{T_i}\}_{i \in [n]}$ by $\boldsymbol{z}^T$; that is, $\boldsymbol{z}^T=\psi(\{\boldsymbol{z}^{T_i}\}_{i \in [n]})$. 

In AT-AKA, we distill the knowledge of the ensemble of teachers into student via logit matching resulting the following loss function for the student model
\begin{align} \label{eq:logitmatching}
 &\mathcal{L}=(1-\alpha)\mathcal{L}_\text{CE}(\boldsymbol{p}^S,\boldsymbol{y}) + \alpha \mathcal{L}_\text{MSE} (\boldsymbol{z}^S,\boldsymbol{z}^T), 
 \end{align}
 where $\alpha$ is a trade-off parameter, and $\mathcal{L}_\text{CE}(\boldsymbol{p}^S,\boldsymbol{y})=-\sum_j \boldsymbol{y}_j \log \boldsymbol{p}^S_j$ with $\boldsymbol{p}^S$ being the probability vector obtained by applying softmax function over the logits $\boldsymbol{z}^S$, and $\boldsymbol{y}$ being the one-hot label vector; and $\mathcal{L}_\text{MSE} (\boldsymbol{z}^S,\boldsymbol{z}^T)=\|  \boldsymbol{z}^S-\boldsymbol{z}^T\|_2^2$. Additionally, the teachers' loss function is the conventional cross-entropy function. After introducing the loss function for both students and teachers in AT-AKA, we propose three different sample-adaptive $\psi(\cdot)$ in the sequel.

\subsection{Adaptive amalgamation functions}

\subsubsection{Naive AT-AKA.}
In AT, the rationale behind feeding adversarial samples into the networks is to solve the inner maximization problem in Eq. \eqref{eq:AT}. Therefore, a naive way of selecting $\psi(\cdot)$ in AT-AKA is 
\begin{align} \label{eq:naive}
\psi(\{\boldsymbol{z}^{T_i}\}_{i \in [n]})= \boldsymbol{z}^{T_k}, ~~ \text{where}~ ~k=\arg\max_j(\mathcal{L}_\text{CE}(\boldsymbol{p}^{T_j},\boldsymbol{y})), 
\end{align}
where $\boldsymbol{p}^{T_j}$ is the output probability vector of $T_j$, for $j \in [n]$.
Although this method greedily picks the logits corresponding to the teacher with the higher loss, it suffers from generalization of different types of attacks. To elucidate, the student trained in this manner might be very robust against some specific strong attacks, but not all of them. To this end, in the following we propose some other methods of logits amalgamation. 
\subsubsection{Linear AT-AKA.}
In this method, the resulting $\boldsymbol{z}^T$ is the weighted average of the logit vectors $\{\boldsymbol{z}^{T_i}\}_{i \in [n]}$:
\small
\begin{align}
\psi(\{\boldsymbol{z}^{T_i}\}_{i \in [n]})=\sum_{i=1}^n \lambda_i \boldsymbol{z}^{T_i}, ~~~ 
\text{with}~~ \lambda_i=\frac{\mathcal{L}_\text{CE}(\boldsymbol{p}^{T_i},\boldsymbol{y})}{\sum_{j=1}^n \mathcal{L}_\text{CE}(\boldsymbol{p}^{T_j},\boldsymbol{y})}.
\end{align}
\normalsize
As seen, this method of logit amalgamation, puts more weight over the logit whose respective loss functions have higher values.
\subsubsection{Soft AT-AKA.}
This method of logit combining is similar to the previous method (linear amalgamation). Yet, compared to the linear amalgamation, it puts more weight on the logit of networks with higher loss values. Mathematically, 
\small
\begin{align}
\psi(\{\boldsymbol{z}^{T_i}\}_{i \in [n]})=\sum_{i=1}^n \lambda_i \boldsymbol{z}^{T_i}, 
~~~ \text{with} ~~ \lambda_i=\frac{e^{\beta\mathcal{L}_\text{CE}(\boldsymbol{p}^{T_i},\boldsymbol{y})}}{\sum_{j=1}^n e^{\beta \mathcal{L}_\text{CE}(\boldsymbol{p}^{T_j},\boldsymbol{y})}},
\end{align}
\normalsize
where a higher $\beta$ assigns more weights to the logit corresponding to the teachers with larger loss values (in our experiments, we set $\beta=1$). 
\subsection{Pareto-Optimal AT-AKA.}
Despite the previous three amalgamation methods which were applied in the logit domain, Pareto-Optimal AT-AKA finds an appropriate amalgamation in the soft probability domain. Thus, the input to the $\psi(\cdot)$ function is $n$-tuple of soft probability vectors of teachers $\{\boldsymbol{p}^{T_i}\}_{i \in [n]}$.

The idea of this method of amalgamation is inspired from \cite{du2020agree}. As discussed in \cite{du2020agree}, there might be conflicts and competitions among the knowledge distilled from the ensemble of teachers. As a remedy, they regard the knowledge amalgamation as a multi-objective optimization, and deploy multiple-gradient descent algorithm to find a Pareto optimal solution which resolves these conflicts \cite{10381881,hamidi2024adafed}. Applying this method to AT-AKA, we first solve the following minimization problem
\small
\begin{align} \label{eq:pareto}
\boldsymbol{\lambda}^*=\min_{\boldsymbol{\lambda}}\|\boldsymbol{p}^S-\sum_{i=1}^n \lambda_i \boldsymbol{p}^{T_i}\|^2, ~~~~~~\text{s.t.} ~ \sum_{i=1}^n \lambda_i=1, ~ 0\leq \boldsymbol{\lambda}. 
\end{align}
\normalsize

Note that in Eq. \eqref{eq:pareto} MSE is applied over the probability vectors which is in contrast with conventional KD where either (i) MSE is applied over the logits, or (ii) KL is used to minimized the distance between the probability vectors. 

After finding the optimal $\boldsymbol{\lambda}^*$ from Eq. \eqref{eq:pareto}, we define $\boldsymbol{p}^T=\sum_{i=1}^n \lambda^*_i \boldsymbol{p}^{T_i}$. Then, we alter the loss function in Eq. \eqref{eq:logitmatching} as follows
\begin{align} \label{eq:paretoloss}
\mathcal{L}=(1-\alpha)\mathcal{L}_\text{CE}(\boldsymbol{p}^S,\boldsymbol{y}) + \alpha \mathcal{L}_\text{KL} (\boldsymbol{p}^S,\boldsymbol{p}^T).    
\end{align}
In the following section, we introduce another variant of AT-AKA, namely collaborative AT-AKA. 
\section{Collaborative AT-AKA}
The goal of AT-AKA was to exploit the capability of cumbersome teachers in learning robust classification boundaries; and then, distill this knowledge into the student of smaller size to yield a small model with high robustness. Concisely speaking, AT-AKA wanted to train a (i) robust and (ii) small networks.

\begin{figure}[!t]
\centering\includegraphics[width=0.7\linewidth]{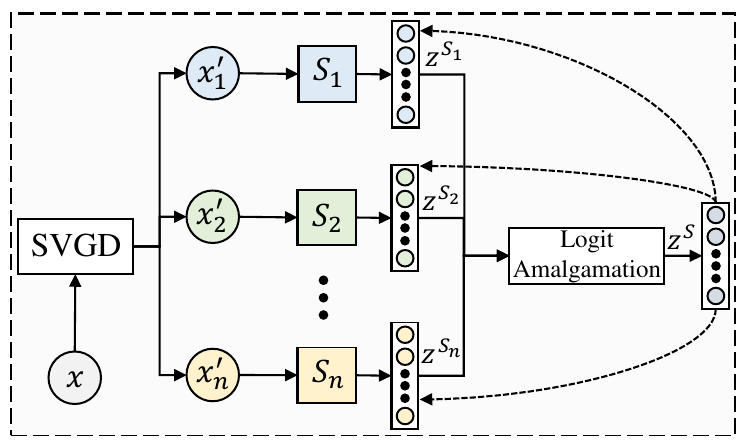}
\caption{The CAT-AKA framework. All the models are students, denoted by $\{S_i\}_{i \in [n]}$. All the students are trained from scratch. The clean sample $\boldsymbol{x}$ is fed to the SVGD block to generate $n$ distinct adversarial samples. Then, the logits of the students are amalgamated, and the resulting logit is used as a supervision for the students.} 
\vskip -0.2in
\label{fig:CAT-AKA}
\end{figure}

In this section, we only care about training a robust model, and not necessarily a small one. To this end, we incorporate online collaborative knowledge distillation \cite{guo2020online} into our framework yielding Collaborative AT-AKA (CAT-AKA). The aim is to feed the students with distinct adversarial examples such that they collaboratively learn a good generalization against adversarial samples. In CAT-AKA, all the models are students of the same size, and the supervision is provided by combining
the output (logits) of these students (See Fig. \ref{fig:CAT-AKA}). Similarly to AT-AKA, all the models would be trained at the same time in CAT-AKA. However, the difference with AT-AKA is that once the training is over, each student can predict independently, and hence the student with the best robustness can be selected. 

\subsection{CAT-AKA formulation}
Denote by $\boldsymbol{p}^{S_i}$ and $\boldsymbol{z}^{S_i}$ the output probability and logit vectors of $S_i$, respectively, for $i \in [n]$. Then, we use the three different logit amalgamation functions $\psi(\cdot)$ discussed earlier to combine the logits of students yielding $\boldsymbol{z}^S=\psi(\{\boldsymbol{z}^{S_i}\}_{i \in [n]})$. Thereafter, the common loss function for each student is defined as  
\begin{align} \label{eq:CATloss}
 &\mathcal{L}=\sum_{i=1}^n(1-\alpha)\mathcal{L}_\text{CE}(\boldsymbol{p}^{S_i},\boldsymbol{y}) + \alpha \mathcal{L}_\text{MSE} (\boldsymbol{z}^{S_i},\boldsymbol{z}^S).
\end{align}
\section{Experiments
} \label{Sec:Experiments}
\begin{table*}[t]
\caption{Adversarial robustness accuracy (\%) on CIFAR-$\{10,100\}$ datasets for AT-AKA variants and some benchmark methods. ResNet-18 and MobileNetV2 are denoted by RN-18 and MN, respectively. $\|\boldsymbol{\epsilon}\|_{\infty}<8/255$. The best results are blodfaced.}
\vskip -0.3in
\label{tablecifar10}
\begin{center}
\begin{small}
\resizebox{\textwidth}{!}{%
\begin{tabular}{c|c|l|cccc||c|l|cccc}
\toprule
Dataset & Model & Method & Clean & FGSM & PGD & AA & Model & Method & Clean & FGSM & PGD & AA\\
\midrule \midrule
\parbox[t]{5mm}{\multirow{9}{3em}{\rotatebox[origin=c]{90}{CIFAR-10}}}& \multirow{9}{3em}{RN-18} & Natural & \textbf{94.60} & 19.11 & 0.0 & 0.0 & \multirow{9}{2em}{MN} & Natural & \textbf{92.90} & 14.53 & 0.0 & 0.0 \\
&& en-SAT &  84.35 &  57.33 & 52.51& 46.80 & & en-SAT &  83.40 &  57.54 & 52.50 & 47.42\\
&& en-ARD &  85.00 &  60.27 & 55.51 & 50.10 && en-ARD &  84.22 &  59.16 & 53.59 & 49.33 \\
 && en-IAD &  84.28 &  59.52 & 55.25 & 50.07 &&  en-IAD &  82.95 &  58.10 & 54.48 & 49.33\\
 && en-RSLAD &  84.45 &  60.95 & 57.00 & 52.52 && en-RSLAD &  84.20 &  60.26 & 55.77 & 51.17\\
 && (N)AT-AKA &  84.28 &  61.15 & 62.07 & 53.02 && (N)AT-AKA & 83.80 &  63.12 &   61.12 & 49.19\\
 && (L)AT-AKA &  84.18 &  65.12 & 63.12 & 54.77 && (L)AT-AKA &  83.72 &  64.12 & 63.77 &52.81\\
 && (S)AT-AKA &  84.10 &  \textbf{65.88} & \textbf{64.20} & \textbf{54.80}&& (S)AT-AKA &  84.00 &  66.44 & \textbf{64.34} & \textbf{54.49}\\
 && (P)AT-AKA &  84.24 &  64.25 & 62.73 & 53.71 && (P)AT-AKA &  84.07 &  \textbf{67.00} & 63.02 & 52.91 \\
\bottomrule
\bottomrule
\parbox[t]{5mm}{\multirow{9}{3em}{\rotatebox[origin=c]{90}{CIFAR-100}}}& \multirow{9}{3em}{RN-18} & Natural & \textbf{75.45} & 10.01 & 0.0 & 0.0 & \multirow{9}{2em}{MN} & Natural & \textbf{74.50} &  7.19 & 0.0 & 0.0 \\
&& en-SAT     & 58.36 &  29.53 & 26.51 & 22.22 &&   en-SAT &  57.81 &  32.95 & 30.50 & 25.60\\
&& en-ARD    &  61.55 &  34.35 & 31.21 & 26.69 &&   en-ARD &  60.73 &  34.15 & 31.28 & 26.43 \\
 && en-IAD   &  58.88 &  34.12 & 31.71 & 26.31 &&   en-IAD &  57.22 &  33.81 & 31.71 & 26.71\\
 && en-RSLAD &  58.70 &  35.21 & 32.80 & 27.65 && en-RSLAD &  59.90 &  35.03 & 32.36 & 27.12\\
 && (N)AT-AKA &  58.90 &  36.41 & 33.40 & 28.12 && (N)AT-AKA &  58.22 &  38.12 & 33.25 &27.50\\
 && (L)AT-AKA &  58.75 &  38.01 & 36.12 & 28.81 && (L)AT-AKA &  58.12 &  38.54 & 35.41 & 29.0\\
 && (S)AT-AKA &  58.65 &  \textbf{40.81} & \textbf{38.10} & \textbf{31.71} && (S)AT-AKA &  58.02 &  \textbf{40.01} & \textbf{38.91} & \textbf{30.94}\\
 && (P)AT-AKA &  58.71 &  38.51 & 37.90 & 30.55 && (P)AT-AKA &  58.76 &  39.00 & 37.49 & 29.99\\
 \bottomrule
\end{tabular}}
\end{small}
\end{center}
\vskip -0.1in
\end{table*}

In this section, we conclude the paper with several experiments to demonstrate the performance of AT-AKA (and CAT-AKA), and compare its effectiveness with state-of-the-art alternatives under some performance
metrics, including clean (natural) accuracy and adversarial accuracy. In Subsection \ref{sub:Experiments}, we analyze the performance of AT-AKA. We first introduce the experimental settings
and implementation details. Then, we conduct a series of
experiments to evaluate the performances of student models obtained by different distillation methods and the transferability of their learnt classification boundaries. Next, in Subsection \ref{sec:ablation} we conduct ablation study on AT-AKA to show that SVGD is a necessary component in AT-AKA framework. Lastly, in Subsection \ref{sub:CAT-AKA} we evaluate the effectiveness of CAT-AKA against some known benchmark methods. The link to the source code for this paper is provided in the \textit{Supplementary materials}. 

\subsection{Experiments on AT-AKA} \label{sub:Experiments}
\noindent $\bullet$ \textbf{Benchmark methods.}
To ensure a fair comparison, for the benchmark methods, we  deploy an ensemble model with the same number of sub-models (teachers in case of adversarial robust distillation). The benchmark methods are (i) natural training method; (ii) one AT method, namely SAT \cite{madry2017towards}; and (iii) three adversarial robust distillation methods, namely ARD \cite{goldblum2020adversarially}, IAD \cite{zhu2021reliable} and RSLAD \cite{zi2021revisiting}.
 
\noindent $\bullet$ \textbf{Datasets.} 
We use CIFAR-$\{10,100\}$ datasets \cite{krizhevsky2009learning} which are commonly used in adversarial robustness. 

\noindent $\bullet$ \textbf{Teacher and student models.} 
For the student models (and for both CIFAR-10 and CIFAR-100), we use two different architectures, namely ResNet-18 \cite{he2016deep} and MobileNetV2 \cite{sandler2018mobilenetv2}. For the teacher models, we use the same models as those used in RSLAD \cite{zi2021revisiting}; that is, we use WideResNet-34-10 \cite{zagoruyko2016wide} for CIFAR-10 and WideResNet-70-16 \cite{gowal2020uncovering} for CIFAR-100. In all experiments we use an ensemble of three teacher models.  

\noindent $\bullet$ \textbf{Training setup.} Elaborated in \textit{Supplementary materials}.  

\noindent $\bullet$ \textbf{Attacks.} We evaluate the model against some commonly-used adversarial attacks: FGSM \cite{goodfellow2014explaining}, PGD\textsubscript{TRADES}\footnote{PGD\textsubscript{TRADES} differs from the original PGD attack introduced in \cite{madry2017towards} in its step-size.} \cite{zhang2019theoretically}, and AutoAttack (AA) \cite{croce2020reliable}. The
perturbation steps for PGD\textsubscript{TRADES} is 20.

\begin{table}[t]
\caption{Ablation study, ResNet-18 on CIFAR-10.}
\vskip -0.3in
\label{table:ablation1}
\begin{center}
\resizebox{0.5\textwidth}{!}{%
\begin{tabular}{l|cccc}
\toprule
Algorithm & Clean & FGSM & PGD & AA \\
\midrule
(N)AT-AKA with PGD   & 84.35 &  58.54 & 54.44  &  48.28\\
(L)AT-AKA with PGD   & 84.36 &  59.27 & 55.71  & 50.02 \\
(S)AT-AKA with PGD   & 84.25 &  59.37 & 57.02 & 50.51 \\
(P)AT-AKA with PGD   & 84.28 &  58.43 & 60.18 & 50.04 \\
\bottomrule
(N)RSLAD   & 84.32 &  60.54 & 57.34 & 52.17\\
(L)RSLAD   & 84.43 &  61.27 & 57.57 & 52.62 \\
(S)RSLAD   & 84.32 &  60.32 & 57.62 & 52.27 \\
(P)RSLAD   & 84.12 &  59.83 & 57.19 & 52.17 \\
\bottomrule
\end{tabular}}
\end{center}
\vskip -0.3in
\end{table}

\noindent $\bullet$ \textbf{Terminology.}
In the tables, we use "Natural" to indicate conventional training with no adversarial samples. Also, "Clean" indicates the test accuracy of the trained model against clean samples. We use the prefix "en" for the benchmarks methods to indicate that they also deploy ensemble method. For brevity, in the tables, we refer to Naive ATA-AKA, Linear ATA-AKA, Soft ATA-AKA and Pareto-optimal ATA-AKA by (N)ATA-AKA, (L)ATA-AKA, (S)ATA-AKA and (P)ATA-AKA, respectively. 

\noindent $\bullet$ \textbf{Observations.} 
As seen in Table \ref{tablecifar10}, the adversarial robustness for AT-AKA is \textit{significantly} higher than that of the benchmark methods. In addition, we observe that Soft ATA-AKA yields the highest adversarial accuracy in most cases. It is worth-noting that strong robust methods come at the cost of a bit sacrifice in the clean accuracy. We observe that the amount of drop in the clean accuracy is negligible in the AT-AKA.


\subsection{Ablation study}\label{sec:ablation}
In this subsection, we perform two different ablation studies where the student is ResNet-18 trained on CIFAR-10. The setup is the same as that we used for generating the results of Table \ref{tablecifar10} (furthermore, additional studies on the effect of the number of teachers on the performance of AT-AKA is reported in supplementary materials).

\noindent $\bullet$ \textbf{Ablation 1: Demonstrating the Necessity of SVGD.}
To demonstrate the necessity of using SVGD method for generating adversarial samples in the AT-AKA framework, we instead use PGD attack with different initializations to generate adversarial samples. As seen in Table \ref{table:ablation1}, the adversarial accuracy of AT-AKA significantly drops when the SVGD-generated samples are replaced by those generated using PGD.  

\noindent $\bullet$ \textbf{Ablation 2: en-RSLAD using our amalgamation methods.} Now, we show that deploying our four amalgamation methods over en-RSLAD (instead of a typical averaging in ensemble method), cannot reach the same accuracy as that obtained using AT-AKA. The results are reported in Table \ref{table:ablation1}.

\subsection{CAT-AKA}\label{sub:CAT-AKA}
Lastly, we evaluate the effectiveness of CAT-AKA. We use three similar students to collaboratively train a robust model, where we use CAT-AKA for ResNet-18 and WideResNet-34-10 networks. The training setup is elaborated in supplementary materials.

We compare the performance of CAT-AKA with three benchmarks, namely SAT, TRADE \cite{zhang2019theoretically} and MART \cite{wang2020improving}. We deploy an ensemble of sub-models for the benchmark methods for a fair comparison. Also, in addition to FGSM and PGD attacks, we further use two strong attacks, namely CW \cite{carlini2017towards} and Fog \cite{kang2019testing} attacks. We only report the results for CIFAR-$\{10,100\}$ in Table \ref{table2cifar10}. Here, we only report the results for Soft CAT-AKA. Also, since all the students are trained at the same time in CAT-AKA, we report the test accuracy for all of them. To this aim, we use the notation CAT-AKA-$S_i$ to indicate the student number $i$. As seen, CAT-AKA outperforms the other benchmark results in the sense that it yields a higher adversarial accuracy. Furthermore, we observe that although there is no difference among the students during the training, the performance of the students are slightly different.

\begin{table*}[t]
\caption{Adversarial robustness accuracy (\%) on CIFAR-$\{10,100\}$ dataset for CAT-AKA variants and some benchmark methods. ResNet-18 and WideResNet-34-10 are denoted by RN and WR, respectively.}
\vskip -0.3in
\label{table2cifar10}
\begin{center}
\begin{small}
\resizebox{\textwidth}{!}{%
\begin{tabular}{c|l|ccccc||c|l|ccccc}
\toprule
Model & Method & clean & FGSM & PGD & CW & Fog & Model & Method & clean & FGSM & PGD & CW & Fog\\
\midrule
\multirow{6}{1em}{\rotatebox[origin=c]{90}{\textcolor{orange}{CIFAR-10}-RN}} & en-SAT & \textbf{85.21} & 58.03 & 50.52 & 50.47 & 41.59 & \multirow{6}{2em}{\rotatebox[origin=c]{90}{\textcolor{orange}{CIFAR-10}-WR}} & en-SAT & \textbf{87.50} &  60.06 & 51.72 & 52.88 & 43.88 \\
& en-TRADES    &  82.48 &  57.62 & 53.26 & 50.92 & 39.48 &&   en-TRADES &  85.92 &  61.87 & 56.58 & 55.36 & 46.73 \\
 & en-MART   &  84.07 &  61.21 & 54.47 & 51.03 & 42.80 &&  en-MART &  84.62 &  62.61 & 57.49 & 54.28 & 44.34\\
 & (S)CAT-AKA-$S_1$&  82.25 &  \textbf{63.88} & \textbf{60.03} & \textbf{53.12}& \textbf{46.79} && (S)CAT-AKA-$S_1$ &  85.02 &  \textbf{65.41} & \textbf{59.96} & \textbf{56.38} & \textbf{49.40}\\
& (S)CAT-AKA-$S_2$ &  82.99 &  62.18 & 58.42 & 51.43& 44.24 && (S)CAT-AKA-$S_2$ &  85.32 &  65.29 & 59.88 & 56.18& 49.09 \\
& (S)CAT-AKA-$S_3$ &  82.45 &  63.51 & 59.81 & 52.98 & 45.51&& (S)CAT-AKA-$S_3$ &  85.48 &  64.81 & 59.80 & 56.49 & 49.01\\
\bottomrule
\midrule
\multirow{6}{1em}{\rotatebox[origin=c]{90}{\textcolor{green}{CIFAR-100}-RN}} & en-SAT & \textbf{58.12} & 30.10 & 25.70 & 25.40 & 16.79 & \multirow{6}{2em}{\rotatebox[origin=c]{90}{\textcolor{green}{CIFAR-100}-WR}} & en-SAT & \textbf{61.54} &  31.86 & 26.72 & 27.88 & 18.08 \\
& en-TRADES    &  54.88 &  31.31 & 29.26 & 25.92 & 15.48 &&   en-TRADES &  58.91 &  33.67 & 31.52 & 28.47 & 19.77 \\
 & en-MART   &  55.38 &  35.22  & 33.03 & 28.80 & 21.47 &&  en-MART &  59.61 &  37.52 & 33.89 & 30.78 & 23.11\\
 & (S)CAT-AKA-$S_1$&  55.25 &  \textbf{37.60} & \textbf{34.43} & \textbf{31.44}& \textbf{25.79} && (S)CAT-AKA-$S_1$ &  59.42 &  \textbf{38.11} & 35.38 & \textbf{34.40} & \textbf{27.40}\\
& (S)CAT-AKA-$S_2$ &  55.59 &  35.27 & 34.12 & 29.43& 22.54 && (S)CAT-AKA-$S_2$ &  59.38 &  38.10 & \textbf{35.39} & 32.28& 26.14 \\
& (S)CAT-AKA-$S_3$ &  55.58 &  35.11 & 32.86 & 29.08 & 22.22 && (S)CAT-AKA-$S_3$ &  59.42 &  37.86 & 33.97 & 31.12 & 25.05\\
\bottomrule
\end{tabular}}
\end{small}
\end{center}
\vskip -0.2in
\end{table*}

\section{Conclusion}
This paper proposed \textbf{a}dversarial \textbf{t}raining via \textbf{a}daptive \textbf{k}nowledge \textbf{a}malgamation of an ensemble of teachers (AT-AKA). The goal of AT-AKA is to address two well-known shortcomings of AT methods, namely: (i) challenges in training small DNNs, and (ii) vulnerability against unseen types of attacks. Inspired by KD, to realize AT-AKA, we generated a diverse/distinguished set of adversarial samples as the inputs to an ensemble of teachers; and then, we adaptively amalgamated the logtis of these teachers to train a generalized-robust student. Additionally, we incorporated AT-AKA into collaborate knowledge distillation yielding a new form of AT-AKA, referred to as CAT-AKA. Lastly, the effectiveness of AT-AKA over existing AT and adversarial robustness distillation methods was justified via conducting a comprehensive set of experiments.

\clearpage

\bibliographystyle{splncs04}
\bibliography{egbib}

\clearpage

\appendix

\section{Training setups}
\noindent $\bullet$ \textbf{Training setup for Table 2 in the main paper.} Models are implemented in PyTorch and trained on eight RTX 2080 Ti GPUs. The networks are trained by stochastic gradient descent (SGD) optimizer using initial learning rate
0.1, momentum 0.9 and weight-decay 2e-4, and also we set the batch size equal to 128. We set the total number of
training epochs to 300, and the learning rate is divided by
10 at the 200-th, 250-th and 270-th epoch. On the other hand, for natural training, we train the networks for 100 epochs on clean images with standard data augmentations and the learning rate is divided by 10 at the 75-th and 90-th epochs. For the benchmark methods we follow their original settings. Adversarial budget is bounded by its $l_{\infty}$ norm, such that $\|\boldsymbol{\epsilon}\|_{\infty}<8/255$. In all the experiments, $\alpha$ is greedily tuned and the best result in terms of accuracy is reported in the tables.

For SVGD method, we use Gaussian RBF kernel: $k(x,x^{\prime})=\text{exp}\{ \frac{-\|x-x^{\prime} \|^2}{2\sigma^2}\}$ with $\sigma=1e-3$, and the number of particles/samples is the same as that of teachers. Also, $\eta=0.003$, and the number of iterations is 20.

\noindent $\bullet$ \textbf{Training setup for Table 4 in the main paper.} Both ResNet-18 and WideResNet-34-10 networks
are trained via SGD optimizer of batch size equals 128 for
120 epochs. For ResNet-18, the initial learning rate is set to 0.01,
momentum to 0.9, and weight decay to 3.5e-3. Also, for WRN-34-10, we set the initial learning rate to 0.1,
momentum to 0.9, and weight decay to 7e-4. The learning rate is divided by 10 at the 75-th, 90-th and 100-th epochs. All the other settings, including those for SVGD and attacks, are the same as those we used for Table 2. In addition, $\|\boldsymbol{\epsilon}\|_{\infty}<8/255$.

\section{Effect of number of teachers}
In this section, we evaluate the effect of the number of teachers over the AT-AKA's performance. To this aim, we repeat the experiments of Table 2 in the main paper, using different number of teachers $n \in \{1,2,3,4\}$. We only report the results for (S)AT-AKA for which we obtained the best results among the other AT-AKA variants. Furthermore, we use the notation (S)AT-AKA$_i$, $i \in \{1,2,3,4\}$, to indicate that $i$ teachers are used for the underlying method. All the training setups are similar to what we used for our experiments in the main paper. 

As seen in Table \ref{tablecifar10} (the next page), AT-AKA does not perform well when we use only one teacher. The reason is that the attacks generated by SVGD are not as strong as those generated by PGD (for the comparison of the attacks generated by PGD and SVGD please refer to the Table 1 in the main paper). As the number of teachers increase, the robustness becomes higher; because the SVGD can successfully generate a diverse set of attacks. Based on the results in Table \ref{tablecifar10}, we can say that if the number of teachers is equal to three, we reach a good trade-off between the complexity and the robustness; this is why we always use three teachers for our analysis in the main paper.

\begin{table*}[!t]
\caption{Effect of the number of Teachers in AT-AKA: adversarial robustness accuracy (\%) on CIFAR-$\{10,100\}$ datasets for AT-AKA variants and some benchmark methods. ResNet-18 and MobileNetV2 are denoted by RN-18 and MN, respectively. $\|\boldsymbol{\epsilon}\|_{\infty}<8/255$. The best results are blodfaced.}
\vskip -0.3in
\label{tablecifar10}
\begin{center}
\begin{small}
\resizebox{0.9\textwidth}{!}{%
\begin{tabular}{c|c|l|cccc||c|l|cccc}
\toprule
Dataset & Model & Method & Clean & FGSM & PGD & AA & Model & Method & Clean & FGSM & PGD & AA\\
\midrule \midrule
\parbox[t]{5mm}{\multirow{5}{3em}{\rotatebox[origin=c]{90}{CIFAR-10}}}& \multirow{5}{3em}{RN-18} & Natural & 94.60 & 19.11 & 0.0 & 0.0 & \multirow{5}{2em}{MN} & Natural & 92.90 & 14.53 & 0.0 & 0.0 \\
&& (S)AT-AKA$_1$ &  84.43 &  55.15 & 51.09 & 42.02 && (S)AT-AKA$_1$ & 84.01 &  55.12 &   51.12 & 45.19\\
 && (S)AT-AKA$_2$ &  84.22 &  59.81 & 58.32 & 50.49 && (S)AT-AKA$_2$ &  83.25 &  60.18 & 58.32 &51.81\\
 && (S)AT-AKA$_3$ &  84.10 &  65.88 & 64.20 & 54.80 && (S)AT-AKA$_3$ &  84.00 &  66.44 & 64.34 & 54.49\\
 && (S)AT-AKA$_4$ &  84.24 &  65.90 & 64.33 & 54.90 && (S)AT-AKA$_4$ &  84.12 &  66.54 & 64.42 & 54.70 \\
\bottomrule
\bottomrule
\parbox[t]{5mm}{\multirow{5}{3em}{\rotatebox[origin=c]{90}{CIFAR-100}}}& \multirow{5}{3em}{RN-18} & Natural & 75.45 & 10.01 & 0.0 & 0.0 & \multirow{5}{2em}{MN} & Natural & 74.50 &  7.19 & 0.0 & 0.0 \\
 && (S)AT-AKA$_1$ &  58.99 &  27.41 & 24.42 & 20.32 && (S)AT-AKA$_1$ &  58.47 &  30.41 & 27.25 &21.51\\
 && (S)AT-AKA$_2$ &  58.85 &  35.01 & 33.51 & 27.72 && (S)AT-AKA$_2$ &  58.33 &  37.59 & 32.35 & 27.42\\
 && (S)AT-AKA$_3$ &  58.65 &  40.81 & 38.10 & 31.71 && (S)AT-AKA$_3$ &  58.02 &  40.01 & 38.91 & 30.94\\
 && (S)AT-AKA$_4$ &  58.52 &  40.91 & 38.25 & 31.79 && (S)AT-AKA$_4$ &  57.92 &  40.18 & 39.07 & 30.99\\
 \bottomrule
\end{tabular}}
\end{small}
\end{center}
\vskip -0.1in
\end{table*}

\end{document}